\pdfoutput=1

\documentclass[11pt]{article}

\usepackage[preprint]{acl}

\usepackage{times}
\usepackage{latexsym}

\usepackage[T1]{fontenc}

\usepackage[utf8]{inputenc}

\usepackage{microtype}

\usepackage{inconsolata}

\usepackage{hyperref}
\usepackage{url}
\usepackage{multirow}
\usepackage{booktabs}
\usepackage{graphicx}
\usepackage{dsfont}
\usepackage{wrapfig}
\usepackage{subfig}
\usepackage{overpic}
\usepackage{color}
\definecolor{mydarkblue}{RGB}{0, 0, 255}
\usepackage{times}
\usepackage{epsfig}
\usepackage{amsthm, amsmath}
\usepackage{mathrsfs,amssymb}
\usepackage{pifont}
\usepackage{enumitem}
\usepackage{verbatim}
\usepackage{makecell}
\usepackage{xcolor}

\title{Android in the Zoo: Chain-of-Action-Thought for GUI Agents}

\author{
Jiwen Zhang$^{1,2}$\thanks{\, This work was done during this author's internship at Shanghai Research Center of Huawei Inc.}, Jihao Wu$^{2}$, Yihua Teng$^{2}$, Minghui Liao$^2$, \\
\textbf{Nuo Xu$^{2}$, Xiao Xiao$^{2}$, Zhongyu Wei$^1$, Duyu Tang$^{2}$}\\
$^1$Fudan University \qquad $^2$Huawei Inc. \\
\texttt{jiwenzhang21@m.fudan.edu.cn} \\
\texttt{\{wujihao,tengyihua,liaominghui1,xunuo4,xiaoxiao55\}@huawei.com} \\
\texttt{zywei@fudan.edu.cn} \quad \texttt{duyutang@huawei.com} \\
\centerline{\url{https://github.com/IMNearth/CoAT}}
}

\begin{document}
\maketitle

\begin{abstract}
Large language model (LLM) leads to a surge of autonomous GUI agents for smartphone, which completes a task triggered by natural language through predicting a sequence of actions of API.
Even though the task highly relies on past actions and visual observations, existing studies typically consider little semantic information carried out by intermediate screenshots and screen operations. 
To address this, this work presents \textbf{Chain-of-Action-Thought} (dubbed \textbf{CoAT}), which takes the description of the previous actions, the current screen, and more importantly the action thinking of what actions should be performed and the outcomes led by the chosen action. 
We demonstrate that, in a zero-shot setting upon three off-the-shelf LMMs, CoAT significantly improves the action prediction compared to previous proposed context modeling. 
To further facilitate the research in this line, we construct a dataset \textbf{Android-In-The-Zoo} (\textbf{\textsc{AitZ}}), which contains 18,643 screen-action pairs together with chain-of-action-thought annotations. 
Experiments show that fine-tuning a 1B model (i.e. AUTO-UI-base) on our \textbf{\textsc{AitZ}} dataset achieves on-par performance with CogAgent-Chat-18B. 
\end{abstract}

\begin{figure*}[t]
\vspace{-0.3cm}
\setlength{\belowcaptionskip}{0cm}
\setlength{\abovecaptionskip}{0.1cm}
    \centering
    \includegraphics[width=\linewidth]{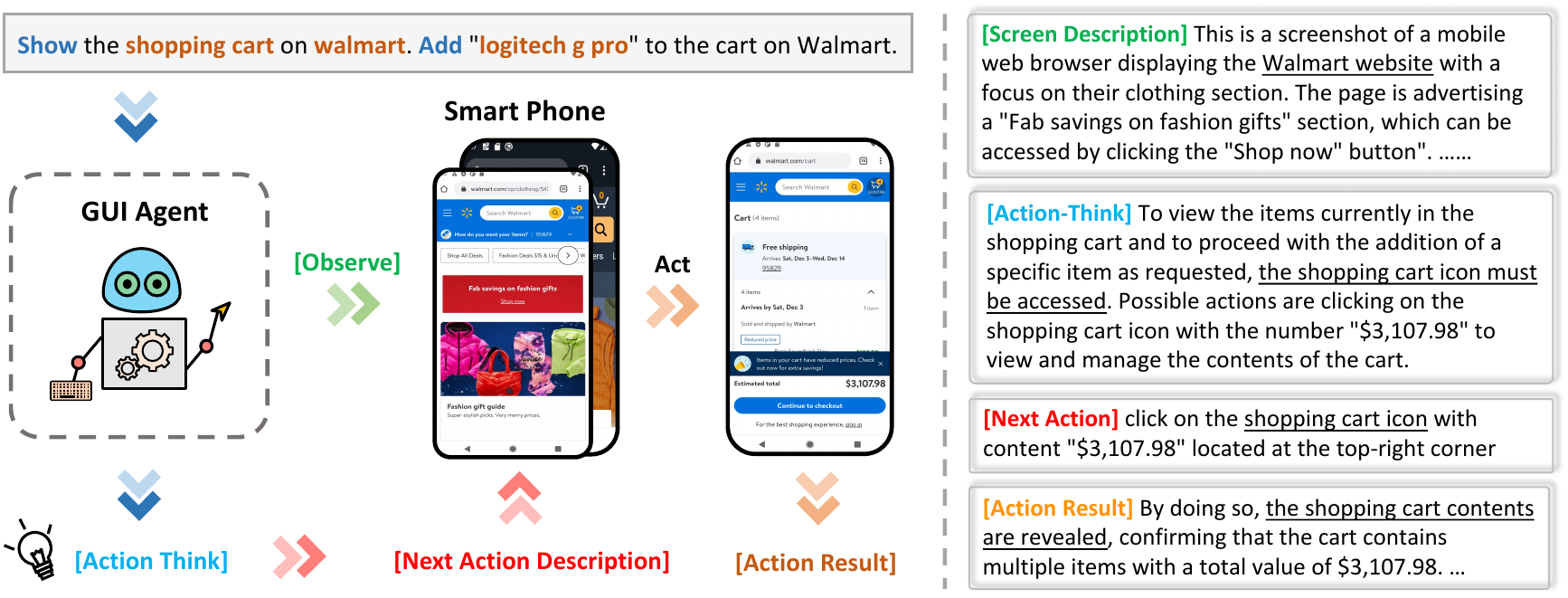}
    \caption{\textbf{The working process of Chain-of-Action-Thought}. The agent will observe the screen, think about actions on current screen to fulfill the user query, describe its next action, act and finally reflect on action results.}
    \label{fig:coat_pipe}
    \vspace{-0.2cm}
\end{figure*}

\begin{table*}[t]
\setlength{\abovecaptionskip}{0.15cm}
\setlength{\belowcaptionskip}{0cm}
\centering
\resizebox{1\textwidth}{!}{
\begin{tabular}{l|c|c|c|c|c|c|c|c|c}
\toprule
\multirow{3}{*}{\textbf{Dataset}}   & \multirow{3}{*}{\textbf{\#Episodes}} & \multirow{3}{*}{\begin{tabular}[c]{@{}c@{}}\textbf{\#Unique}\\ \textbf{Instructions}\end{tabular}} & \multirow{3}{*}{\textbf{\#Apps}} & \multirow{3}{*}{\textbf{\#Steps}} & \multicolumn{5}{c}{\textbf{Annotation}}   \\ \cmidrule{6-10}
  && && & \begin{tabular}[c]{@{}c@{}}\textbf{screen}\\ \textbf{desc}\end{tabular} & \begin{tabular}[c]{@{}c@{}}\textbf{action}\\ \textbf{coord}\end{tabular} & \begin{tabular}[c]{@{}c@{}}\textbf{action}\\ \textbf{desc}\end{tabular} & \begin{tabular}[c]{@{}c@{}}\textbf{action}\\ \textbf{thinking}\end{tabular} & \begin{tabular}[c]{@{}c@{}}\textbf{episode}\\ \textbf{feasibility}\end{tabular} \\ \midrule
PixelHelp~\citep{li2020mapping}   & 187     &  187   &  4    & \textasciitilde 4  &              & $\checkmark$&&&   \\
MoTIF~\citep{burns2021mobile}     & 4707    &  270   & 125   & 4.5                &              & $\checkmark$&&& $\checkmark$ \\
UGIF~\citep{venkatesh2022ugif}    & 523     &  480   & 12    & 6.3                &              & $\checkmark$&$\checkmark$&&   \\
Meta-GUI~\citep{Sun2022METAGUITM} & 4684    & 1125   & 11    & 5.3                &              & $\checkmark$&&&   \\
AITW~\citep{rawles2023android}    & 715142  & 30378  & 357+  & 6.5                &              & $\checkmark$&&&   \\ \midrule
\textbf{\textsc{AitZ}} (Ours)    &  2504   & 2504   & 70+   & 7.5                & $\checkmark$ & $\checkmark$& $\checkmark$ & $\checkmark$  & $\checkmark$ \\
\bottomrule
\end{tabular}
}
\caption{\textbf{Comparison of \textbf{\textsc{AitZ}} to existing Android GUI datasets.} We consider the number of episodes, instructions, related apps, average steps and granularity of annotations. Specifically, action semantics includes action descriptions and action thinkings, while episode feasibility refers to the success verification of collected episodes.}
\label{tab:dataset_card}
\vspace{-0.1cm}
\end{table*}

\section{Introduction}

Nowadays, smartphones have become an essential part of daily lives. Autonomous operation of Graphical User Interfaces (GUI) by human instructions can substantially simplify everyday routines. 
Such tasks, formalized as \textbf{GUI Navigation}~\citep{li2020mapping, sun2022meta}, therefore carry immense social importance, especially for people with physical disabilities~\citep{nanavati2023physically}. 

Recent works have explored prompt engineering~\citep{wen2023empowering, zhan2023you}, finetuning~\citep{hong2023cogagent} and memory augmentation~\citep{lee2023explore} to utilize the capability of large language models (LLM) on interactive mobile environments. 
However, progress is held back due to the scarcity of attention paid on the underlying semantics of smartphone operations. 
GUI navigation usually entails initially observing the screen, considering the next action to take, and reflecting on the outcome of that action~\cite{zhang2024ufo}. 
Previous works~\cite{zhan2023you, cheng2024seeclick} ignore the logic behind diverse actions on the screen, concentrating solely on the coordinates of an operation, such as ``click on (0.17, 0.89)'', which is quite insufficient.
As shown in Figure~\ref{fig:coat_pipe}, we need explicit explanations for the intermediate results during GUI navigation:
\begin{itemize}[topsep=0pt,leftmargin=*]
    \setlength{\itemsep}{1pt}
    \setlength{\parsep}{2pt}
    \setlength{\parskip}{0pt}
    \item \textbf{Screen Context}: In which app or interface did the action occur? This helps to learn the background and possible effects of the action.
    \item \textbf{Action Think}: Why the specific action on the current screen is chosen? Does it facilitate the completion of user query? Such thinking process helps the agent to better capture the user intent.
    \item \textbf{Action Target}: Which UI element is the action operating on? A button, an icon, or a link? 
    \item \textbf{Action Result}: What change will this action cause? Understanding this ensures the consistency of the agent decision-making process.
\end{itemize}
In order to equip existing GUI agents with such capability, we summarize the series of navigation steps as \textbf{Chain-of-Action-Thought} (\textbf{CoAT}), including the screen description, the thinking process about the next action, the textual next action description, and the possible action outcomes. Screen description, together with the screenshots, provides the agent with information basis for decision-making~\cite{wang2021screen2words}. Whereas action think, action description and action result demonstrate the rationale between operations. 
Equipped with CoAT, we achieve significant improvements in the action prediction across three off-the-shelf large multimodal models (LMM) compared to standard context prompting, including GPT-4V~\cite{openai2023gpt4}, Gemini-Pro-Vision~\cite{team2023gemini} and Qwen-VL-Max~\cite{bai2023qwen}. 

However, complex context modeling of language models emerges at a large model scale~\cite{zhang2023multimodal}. Without high quality CoAT-driven data, smaller models can not possess the desired ability through fine-tuning.
To remedy this blank, we propose a new dataset \textbf{Android-In-The-Zoo} (\textbf{\textsc{AitZ}}). 
\textbf{\textsc{AitZ}} is the first dataset that connects the perception (of screen layouts and UI elements) and the cognition (of action decision-making process) together.
Based on the screen episodes from~\citep{rawles2023android}, we leverage the most-capable proprietary model, GPT-4V~\citep{openai2023gpt4}, and state-of-the-art icon detection model~\citep{liu2018learning} to generate candidate answers for the screen descriptions, action thinkings and next action descriptions. 
These candidates are further validated and refined by human to guarantee alignment with the screenshots. Finally, \textbf{\textsc{AitZ}} contains about 19,000 screenshots spanning over 70 Android apps, coupled with 4$\times$ useful annotations compared with action coordinate labels only. We verify the effectiveness of CoAT by additionally finetuning a small multimodal agent from scratch on our \textbf{\textsc{AitZ}} dataset. Experiments show that our proposed chain-of-action-thought improves both the goal progress and the learning efficiency of GUI agents.


Our contributions are summarized as follows:
\begin{itemize}[topsep=0pt,leftmargin=*]
    \setlength{\itemsep}{1pt}
    \setlength{\parsep}{2pt}
    \setlength{\parskip}{0pt}
    \item We propose \textbf{Chain-of-Action-Thought} (\textbf{CoAT}), a novel prompting paradigm to explicitly capture the underlying semantics during navigation actions, allowing GUI agents to perceive, think and decide in an interleaved manner. 
    \item We construct Android-In-The-Zoo (\textbf{\textsc{AitZ}}), the first and largest fine-grained dataset in the Android GUI navigation field. \textbf{\textsc{AitZ}} consisting of 2504 unique instructions and 18,643 screen-action pairs together with four types of semantic annotations, spanning over 70 Android apps.
    \item We conduct both zero-shot and fine-tuning evaluation on the \textbf{\textsc{AitZ}} dataset, validating the necessity and effectiveness of proposed chain-of-action-thought prompting. 
\end{itemize}

\begin{figure*}[t]
\vspace{-0.3cm}
\begin{minipage}{0.59\textwidth}
\centering
\setlength{\belowcaptionskip}{0cm}
\setlength{\abovecaptionskip}{0.1cm}
    \centering
    \includegraphics[width=\linewidth]{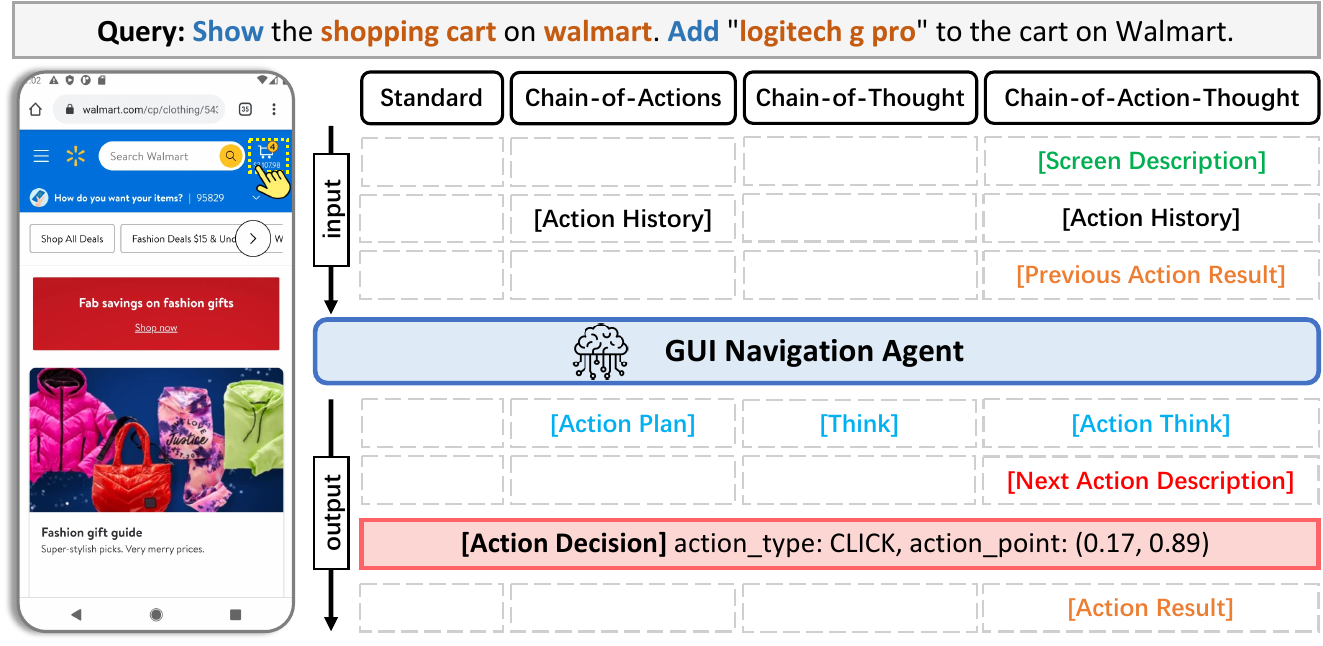}
    \caption{\textbf{Chain-of-Action-Thought compared with three typical prompting methods for GUI tasks,} including Standard~\cite{rawles2023android} prompting, Chain-of-Action~\cite{zhan2023you} prompting and Chain-of-Thought~\cite{wei2022chain} prompting.}
    \label{fig:coat_cmp_other}
    \vspace{-0.0cm}
\end{minipage}
\begin{minipage}{0.03\textwidth}
\end{minipage}
\begin{minipage}{0.38\textwidth}
\centering
    \resizebox{\linewidth}{!}{
    \begin{tabular}{c|c|ccc}
    \toprule
    \multicolumn{1}{l|}{\multirow{2}{*}{\textbf{Prompt}}} & \multicolumn{1}{l|}{\multirow{2}{*}{\textbf{Metric}}} & \multicolumn{3}{c}{\textbf{Model}}                                                               \\ \cmidrule{3-5}
    \multicolumn{1}{l|}{}                        & \multicolumn{1}{l|}{}                        & \multicolumn{1}{l}{QwenVL} & \multicolumn{1}{l}{Gemini-PV} & \multicolumn{1}{l}{GPT-4V} \\ \midrule
    \multirow{2}{*}{\textbf{CoA}}                        & hit                                         & 94.5                       & \textbf{99.8}                          & \underline{99.3}                      \\ \cmidrule{2-5}
                                                & acc                                         & 44.4                       & \underline{47.7}                         & \textbf{62.8}                       \\ \midrule
    \multirow{2}{*}{\textbf{CoT}}                        & hit                                         & 95.6                       & \textbf{97.5}                          & \underline{97.1}                      \\ \cmidrule{2-5}
                                                & acc                                         & 49.4                       & \underline{52.0}                          & \textbf{64.1 }                      \\ \midrule
    \multirow{2}{*}{\textbf{CoAT}}                       & hit                                         & 96.3                       & \underline{96.4}                         & \textbf{98.2}                       \\ \cmidrule{2-5}
                                                & acc                                         & 52.4                       & \underline{54.5}                         & \textbf{73.5 }             \\ \bottomrule        
    \end{tabular}
    }
    \captionof{table}{\textbf{Quantitative comparison of three prompting methods} on Qwen-VL-Max, Gemini-1.0-Pro-Vision and GPT-4V. CoA and CoT are short for chain-of-action and chain-of-thought, respectively. ``hit'' means format hit rate, and ``acc'' means accuracy.}
    \label{tab:coat_cmp_other}
    \vspace{-0.0cm}
\end{minipage}
\vspace{-0.2cm}
\end{figure*}

\section{Chain-of-Action-Thought (CoAT)}

\subsection{Definition}

Consider a general GUI navigation agent with a user query $u \in \mathcal{U}$ to solve. At time step $t$, an agent receives a screenshot observation $o_t \in \mathcal{O}$ from the environment and takes an action $a_t \in \mathcal{A}$ following some policy $\pi(a_t|o_t,h_{t-1},u)$ where $h_{t-1}=(o_1, a_1, ...,o_{t-1}, a_{t-1})$ is the history for the agent. Directly learning the policy is challenging as the relations between history, current observations, and possible actions are highly implicit. For example, knowing the search bar is already active is necessary for an agent to make the next action decision to type text. Therefore, we define Chain-of-Action-Thought (CoAT) as a shortcut to comprehend the interaction dynamics during navigation. 

The basic components of CoAT, marked as grey-bordered boxes on the right side of Figure~\ref{fig:coat_pipe}, are:
\begin{itemize}[topsep=0pt,leftmargin=*]
    \setlength{\itemsep}{1pt}
    \setlength{\parsep}{2pt}
    \setlength{\parskip}{0pt}
    \item \textbf{Screen Description} (SD) describes the main content of the given screenshots, including the screen type and primary apps or widgets presented. Screen description provides the textual context for further decision-making.
    \item \textbf{Action Think} (AT) analyzes the user query and current screen, and combines the history information to infer the possible actions that help to fulfil the target. Mathematically, action think provides a conditional probability $p(AT|o_t,u,h_{t-1})$. If the action think summarizes the current state perfectly and contains reasonable action plans, the decision can be made by calculating $p(a_t|AT)$.
    \item \textbf{Next Action Description} (AD) illustrates the UI element or screen functions being operated, i.e. ``click on the shopping cart icon'' or ``scroll up to open the app drawer''. Action description helps to form a readable action history.
    \item \textbf{Action Result} (AR) connects the current screen $o_t$ and next action $a_t$ to the future observations $o_{t+1}$, by synthesizing the action outcomes after comparing the screenshot before and after the action. Usually, at time step $t$, we combine last action result $AR_{t-1}$ with previous action descriptions to form a continuous and consistent history. 
\end{itemize}
Since each CoAT component carries useful semantics, it is free to combine them according to language models used. Our further experiments will validate the effectiveness and flexibility of the application of proposed CoAT framework.

\subsection{Comparison}
\label{sec:cmp_prompt}

Figure~\ref{fig:coat_cmp_other} compares proposed CoAT with Standard~\cite{rawles2023android}, Chain-of-Action (CoA)~\cite{zhan2023you} and Chain-of-Thought (CoT)~\cite{wei2022chain} prompting methods. The proposed CoAT carries explicitly more semantic information about the screen and actions. To further validate the effectiveness of CoAT, we conduct a preliminary experiment on 50 episodes randomly sampled from AITW~\citep{rawles2023android} dataset. We select three most capable proprietary models, i.e. GPT-4V~\cite{openai2023gpt4}, Gemini-Pro-Vision~\cite{team2023gemini} and Qwen-VL-Max~\cite{bai2023qwen}, to be the GUI agent and apply different prompting methods on them. To ensure an accurate measurement of action prediction accuracy, we use set-of-mark tagging method~\cite{yan2023gpt} to annotate UI elements on screen. As shown in Table~\ref{tab:coat_cmp_other}, agents with CoAT surpass CoA and CoT by a large margin. Moreover, GPT-4V demonstrates optimal performance, making it a good collaborator for subsequent data collection.

\begin{figure*}[t]
\vspace{-0.3cm}
\setlength{\belowcaptionskip}{0cm}
\setlength{\abovecaptionskip}{0.1cm}
    \centering
    \includegraphics[width=0.95\linewidth]{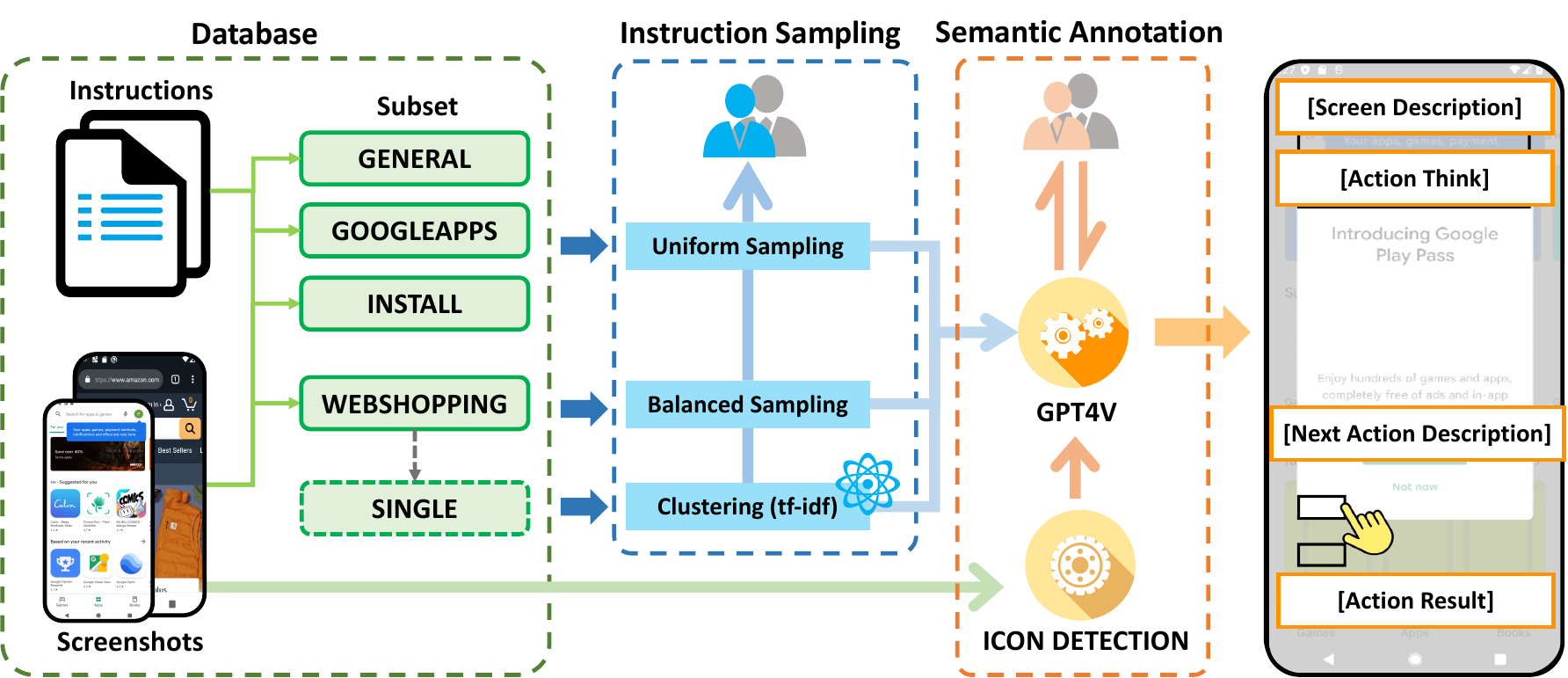}
    \caption{\textbf{\textsc{AitZ} data collection pipeline.} During sampling process, human annotators first verify the clustering results, and then check whether the sampled episode successfully complete the query. During annotation process, human annotators examine and correct the GPT generated semantic descriptions.}
    \label{fig:data_collection}
    \vspace{-0.1cm}
\end{figure*}

\begin{figure*}[t]
\setlength{\belowcaptionskip}{0cm}
\setlength{\abovecaptionskip}{-0.1cm}
    \centering
    \includegraphics[width=\linewidth]{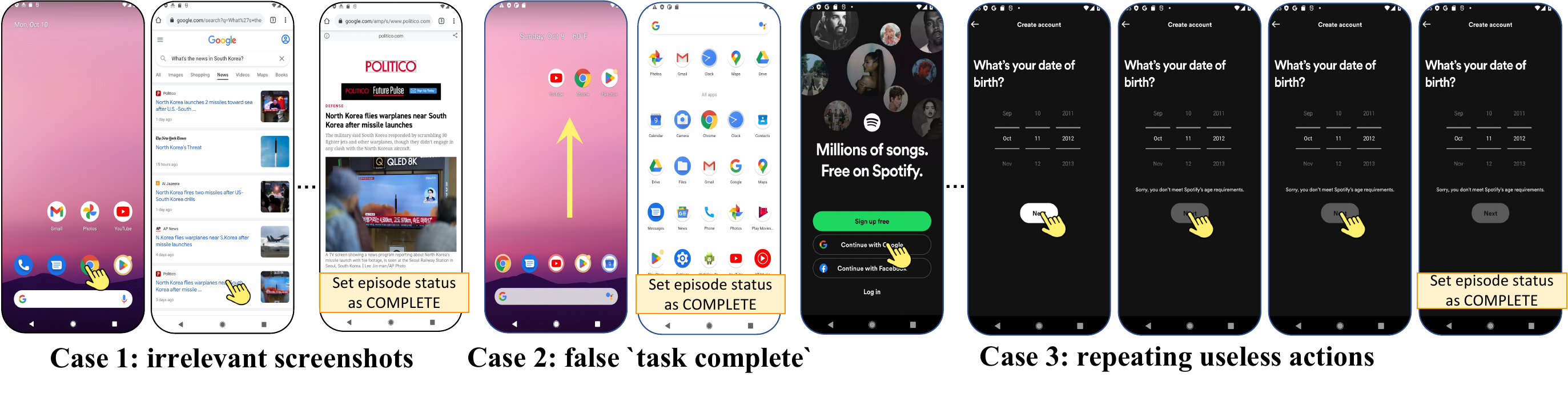}
    \caption{\textbf{Three typical cases of wrong episodes in \textsc{AitW}~\cite{rawles2023android} dataset.} We take the task to `check the settings for the spotify app' as example. There exists 15 episodes corresponding to this instruction, and among them 13 do not actually open the spotify app. This highlights the reasonability to perform data validation.}
    \label{fig:aitw_false}
    \vspace{-0.3cm}
\end{figure*}

\begin{table*}[t]
\vspace{-0.3cm}
\setlength{\abovecaptionskip}{0.15cm}
\setlength{\belowcaptionskip}{0cm}
\centering
\resizebox{1\textwidth}{!}{
\begin{tabular}{c|l|c|c}
\toprule
\textbf{Shopping web/app}        & \multicolumn{1}{c|}{\textbf{Instruction Template}}                                          & \textbf{\#Instructions} & \textbf{\#Episodes} \\ \midrule
\multirow{6}{*}{amazon} & add something to the cart on amazon                                               & 80                            & 180                       \\
                        & clear/empty cart, then add something to the cart on amazon                              & 111                            & 135                        \\
                        & clear/empty cart, search for something, select the first entry and add to cart on amazon & 105                         & 124    \\ 
                        & clear cart, search for something, select the first entry, add to cart on amazon, and checkout   & 110 & 135 \\
                        & show/view the shopping cart, search for something on amazon and add it to the cart & 42 & 52 \\
                        & show/view the shopping cart, add something to the cart on amazon, then checkout  & 59 & 75 \\\bottomrule                   
\end{tabular}}
\caption{\textbf{An example of repeating instructions with the same template on \textsc{webshopping} subset of AITW dataset.} We take instructions related to `amazon' for demonstration. Similar templates can be found for samples related to other shopping websites/apps, including `bestbuy', `ebay', `costco' and `walmart'.}
\label{tab:aitw_repeat_instr}
\vspace{-0.2cm}
\end{table*}

\begin{figure*}[t]
\setlength{\belowcaptionskip}{0cm}
\setlength{\abovecaptionskip}{-0.1cm}
    \centering
    \includegraphics[width=\linewidth]{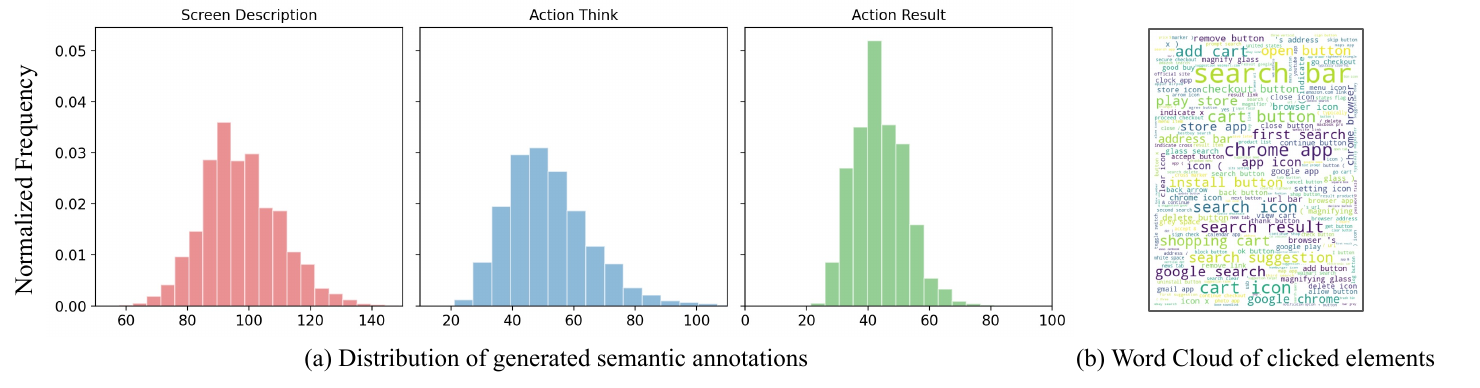}
    \caption{\textbf{Distributions of (a) the length of three different types of semantic annotations and (b) the phrase frequencies of clicked UI elements} on the \textbf{\textsc{AitZ}} dataset. The size of each word corresponds to its tf-idf score.}
    \label{fig:dataset_stat}
    \vspace{-0.2cm}
\end{figure*}

\section{Android in the Zoo (\textbf{\textsc{AitZ}})}

There is a lack of data that captures the underlying semantics of the CoAT paradigm, hindering small models from obtaining this ability. We therefore propose to construct a novel, high-quality and comprehensive dataset to remedy this blank. 

\subsection{Data Collection}

\paragraph{Instruction Sampling} We build our dataset upon the currently most scaled Android GUI navigation dataset, AITW~\citep{rawles2023android}. AITW dataset has 715k episodes spanning 30k unique instructions. We observe that (1) the diversity of instructions mainly comes from the subset \textsc{webshopping}, and these instructions have clear templates, as shown in Table~\ref{tab:aitw_repeat_instr}; (2) the richness of episodes results from subset \textsc{googleapps}, where each instruction corresponds to more than 2000 episodes. 
However, within the AITW dataset, there exists exist numerous mismatch cases between the observed screenshots and the instructions (see Figure~\ref{fig:aitw_false}).
Thus, we sample the instructions and episodes to reduce redundancy and filter the error cases by using a subset-specific sampling strategy:
\begin{itemize}[topsep=0pt,leftmargin=*]
    \setlength{\itemsep}{1pt}
    \setlength{\parsep}{2pt}
    \setlength{\parskip}{0pt}
    \item For subset \textsc{general}, \textsc{googleapps} and \textsc{install}, as there are few unique instructions in each subset, we uniformly sample $x$ samples for each instruction ($x=3,5,3$ respectively).
    \item For subset \textsc{webshopping}, we conduct balanced sampling on the categories of shopping websites/apps and the objects involved.
    \item For subset \textsc{single}, as the instructions are diverse and cluttered, we perform clustering and then conduct balanced sampling on the clustered data. 
\end{itemize}
This results in a total number of 3461 unique instructions, corresponding to 7180 episodes. We recruit ten annotators to manually verify the correctness of the sampled episodes. 
Finally, for 5147 successful episodes, we randomly select one episode paired with each unique instruction. 

\paragraph{Semantic Annotation} It is crucial for GUI agents to understand the screen information and make decisions accordingly. To mitigate the lack of such detailed data, we leverage GPT-4V through Azure-API as the navigation expert and prompt it to do the screen description, action thinking, next action description and action result summarization tasks. Note that the amount of information used to generate semantic annotations varies. For example, the screen description is query-independent, whereas for next action description, both the query and the coordinate of golden actions are provided for reference (see Appendix~\ref{appendix:seman_anno} for more details).  
Thanks to the correctness check at instruction sampling stage, the golden actions have all been verified. We then recruit three experts who have a good understanding of UI elements as annotators to examine whether the generated action description, action thinking and action result match the golden actions. Once inconsistency is found, annotators will manually revise the action descriptions, and enforce GPT-4V to regenerate the action thoughts and action results based on the correct descriptions.

\begin{table}[t]
\setlength{\abovecaptionskip}{0.15cm}
\setlength{\belowcaptionskip}{0cm}
\centering
\resizebox{\linewidth}{!}{
\begin{tabular}{l|cc|cc}
\toprule
\multirow{2}{*}{\textbf{Subset}} & \multicolumn{2}{c|}{\textbf{Train}} & \multicolumn{2}{c}{\textbf{Test}} \\ \cmidrule{2-5}
                        & \textbf{\#Episodes}   & \textbf{\#Screens}  & \textbf{\#Episodes}  & \textbf{\#Screens } \\ \midrule
\textsc{General}                 & 323          & 2405       & 156         & 1202       \\
\textsc{Install}                 & 286          & 2519       & 134         & 1108       \\
\textsc{GoogleApps}              & 166          & 1268       & 76          & 621        \\
\textsc{Single}                  & 844          & 2594       & 0           & 0          \\
\textsc{WebShopping}             & 379          & 5133       & 140         & 1793       \\ \midrule
Total                   & 1998         & 13919      & 506         & 4724      \\ 
\bottomrule
\end{tabular}
}
\caption{\textbf{Detailed statistics of the training and test split of \textsc{AitZ} dataset.} Since \textsc{Single} subset contains single-step tasks only, we place all \textsc{Single} data and related episodes into the training set. }
\label{tab:aitz_split}
\vspace{-0.2cm}
\end{table}

\subsection{Dataset Analysis}

We compare our \textbf{\textsc{AitZ}} dataset with the most related Android GUI navigation datasets, including PixelHelp~\citep{li2020mapping}, MOTIF~\citep{burns2021mobile}, UGIF~\citep{venkatesh2022ugif}, Meta-GUI~\citep{sun2022meta} and \textsc{AitW}~\citep{rawles2023android}. Our dataset contains the same magnitude of human demonstration as these smaller datasets, but with a significantly greater richness of instructions. Table~\ref{tab:dataset_card} demonstrates that our dataset is unique, converting rich semantic information. 

In Figure~\ref{fig:dataset_stat}, we provide statistics of the \textbf{\textsc{AitZ}} dataset, including the distribution of textual lengths and the word cloud of operated UI elements. Specifically, the majority of screen descriptions consist of 80$\sim$120 words, while most action think have 30$\sim$70 words. The action result exhibits a narrower range, from 20 to 80 words. 


\begin{table*}[t]
\vspace{-0.3cm}
\setlength{\abovecaptionskip}{0.15cm}
\setlength{\belowcaptionskip}{0cm}
\centering
\resizebox{1\textwidth}{!}{
\begin{tabular}{c|c|c|cc|cc|c|c|cc|c}
\toprule
\multirow{3}{*}{\textbf{Mode}} & \multirow{3}{*}{\textbf{Model}} & \multicolumn{9}{c|}{\textbf{Atomic}}   & \textbf{Episodic} \\ \cmidrule{3-12}
    &  & \multirow{2}{*}{\textbf{SCROLL}} & \multicolumn{2}{c|}{\textbf{CLICK}} & \multicolumn{2}{c|}{\textbf{TYPE}} & \multirow{2}{*}{\textbf{PRESS}} & \multirow{2}{*}{\textbf{STOP}} & \multicolumn{2}{c|}{\textbf{Total}} & \multirow{2}{*}{\textbf{GP}} \\ \cmidrule{4-7} \cmidrule{10-11}
    &  &   & \textbf{type}   & \textbf{match}   & \textbf{type}   & \textbf{match}  &  & & \textbf{type}   & \textbf{match}   &   \\ \midrule
\multirow{2}{*}{ZS} & \multicolumn{1}{l|}{CogAgent~~~~}  & 56.41 & 79.90    & 51.50 & \textbf{67.40}    & \textbf{34.00}    & \textbf{48.30}    & 4.76    & 65.86    & 44.52 & 13.82 \\ 
    &  \multicolumn{1}{r|}{+CoAT}   & \textbf{70.22} & \textbf{88.23}    & \textbf{66.15} & 45.80    & 21.80    & 45.95    & \textbf{24.60}   & \textbf{72.59}   & \textbf{53.28} & \textbf{17.13} \\ \midrule
\multirow{2}{*}{FT} & \multicolumn{1}{l|}{AUTO-UI~~~~}  & \textbf{74.88} & 44.37    & 12.72 & 73.00    & 67.80    & 49.09    & 60.12   & 73.79    & 34.46 & 6.59  \\
    &  \multicolumn{1}{r|}{+CoAT}    & 61.40 & \textbf{74.56}    & \textbf{32.20} & \textbf{87.80}    & \textbf{81.40}    & \textbf{57.70}    & \textbf{74.40}   & \textbf{82.98}    & \textbf{47.69} & \textbf{14.51}    \\
\bottomrule
\end{tabular}
}
\caption{\textbf{Main results of CogAgent and AUTO-UI on \textbf{\textsc{AitZ}} dataset.} ZS and FT are short for zero-shot and finetuning evaluation, respectively. For CLICK and TYPE actions, which is more complicated than the other three, we additionally report the action type prediction accuracy, marked as `type' in this table. Total action-matching score is also included. `GP' is short for goal progress. The best result of each model is marked in \textbf{bold}.}
\label{tab:main}
\vspace{-0.2cm}
\end{table*}

\section{Experimental Setup}
\label{sec:exp_setup}


\subsection{Baseline Models}

\paragraph{CogAgent~\citep{hong2023cogagent}} is a LLM-based multimodal GUI agent built upon CogVLM~\cite{Wang2023CogVLMVE}. It scales the image resolution up to 1120×1120 by fusing high-resolution features to every decoder layer with cross-attention. CogAgent is pre-trained on a handful of tasks aimed to adapt it for GUI application scenarios, i.e. text recognition~\cite{Schuhmann2022LAION5BAO}, visual grounding~\cite{li2023blip}, and GUI imagery~\cite{hong2023cogagent}. It is further finetuned with GUI tasks on web~\citep{deng2023mind2web} and smartphones~\cite{rawles2023android}. Since the training data for CogAgent is not publicly available, we conduct a zero-shot evaluation to assess to what extent CoAT supports the task. 

\paragraph{AUTO-UI~\citep{zhan2023you}} is a specialized model for GUI navigation on AITW~\citep{rawles2023android} dataset. Screen features are extracted by the encoder from BLIP-2~\citep{li2023blip} and fed into FLAN-Alpaca to decode actions. AUTO-UI is trained on a randomly split training set, covering 80\% of \textsc{AitW} episodes, and evaluated on 10\% randomly selected testing episodes. As stated before, \textsc{AitW} dataset has a large amount of repeating and problematic data, resulting in almost identical distributions between its training and test set. 
Therefore, we train this model from scratch on the training split of \textbf{\textsc{AitZ}} to validate the necessity and helpfulness of the fine-grained semantic annotations provided by \textbf{\textsc{AitZ}} dataset.

\subsection{Evaluation Metrics} 

\paragraph{Atomic Metrics} Following ~\citep{zhan2023you, hong2023cogagent}, we compute the screen-wise action-matching score (``match'' for short). An action is correct if both the action type and the action details (i.e. scroll direction, typed text, clicked position and pressed button) match the gold ones. 

\paragraph{Episodic Metrics} As the GUI navigation is a sequential decision-making problem, it is crucial to evaluate the progress made by the agent towards the user query. Therefore, we propose to use goal progress, a metric indicating the relative position where the first error occurs in the sequence. 

\subsection{Implementation Details}
We randomly split 70\% episodes as training data, and 30\% episodes as testing data (1998/506). It is notable that, as the episodes and instructions in \textbf{\textsc{AitZ}} are distinct, the training set and test set ensure no information leakage. The detailed statistics are in Table~\ref{tab:aitz_split}. For AUTO-UI, we adopt the same weight initialization strategies as ~\cite{zhan2023you} and fine-tune the models up to 10 epochs, with a learning rate of 1e-4. 
For CogAgent, we utilize the trained model weights from CogAgent-Chat and prompt it to use different semantic annotations. 
For both models, we keep the original output format unchanged but add extra information to the input or output of these models.

\begin{table*}[t]
\vspace{-0.4cm}
\setlength{\abovecaptionskip}{0.15cm}
\setlength{\belowcaptionskip}{0cm}
\centering
\resizebox{1\textwidth}{!}{
\begin{tabular}{c|cc|cc|c|cc|cc|c|c|cc|c}
\toprule
&\multicolumn{4}{c|}{\textbf{Semantic Annotations}}  & \multicolumn{9}{c|}{\textbf{Atomic}} & \textbf{Episodic}   \\ \cmidrule{2-15}
&\multicolumn{2}{c|}{\textbf{input}}  & \multicolumn{2}{c|}{\textbf{output}} & \multicolumn{1}{c|}{\multirow{2}{*}{\textbf{SCROLL}}} & \multicolumn{2}{c|}{\textbf{CLICK}}   & \multicolumn{2}{c|}{\textbf{TYPE}} & \multicolumn{1}{c|}{\multirow{2}{*}{\textbf{PRESS}}} & \multicolumn{1}{c|}{\multirow{2}{*}{\textbf{STOP}}} & \multicolumn{2}{c|}{\textbf{Total}} & \multirow{2}{*}{\textbf{GP}} \\ \cmidrule{2-5}\cmidrule{7-10} \cmidrule{13-14}
&\textbf{SD}  & \multicolumn{1}{c|}{\textbf{PAR}} & \textbf{AT}& \textbf{AD}& \multicolumn{1}{c|}{}   & \textbf{type}  & \multicolumn{1}{c|}{\textbf{match}} & \textbf{type}  & \multicolumn{1}{c|}{\textbf{match}}  & \multicolumn{1}{c|}{}  & \multicolumn{1}{c|}{} & \textbf{type}   & \textbf{match}  &    \\ \midrule
(1)& && &  & 74.88   & 44.37  & 12.72   & 73.00  & 67.80& 49.09  & 60.12 & 73.79  & 34.46 & 6.59  \\ \midrule
(2)&$\checkmark$&& &  & \textbf{87.85}   & 49.52  & 20.21   & 81.40  & 64.20& 53.52  & 49.80 & 80.55  & 39.33& 10.71  \\
(3)& & $\checkmark$ & &  & 78.54   & 63.23  & 29.39   & 85.60  & \underline{79.40}& \underline{55.35}  & \textbf{79.17} & \textbf{83.91}  & \textbf{48.35} & \underline{14.06}  \\
(4)&$\checkmark$& $\checkmark$ & &  & 80.53   & 59.10  & 25.95   & 80.60  & 62.40& 55.09  & 57.14 & 81.77  & 42.38& 13.64  \\ \midrule
(5)&   &   & $\checkmark$ &   & \underline{80.87}    & 43.09   & 13.16 & \textbf{89.80}    & 78.60   & 46.74   & 25.00  & 73.45 & 32.68 & 9.08   \\
(6)& && & $\checkmark$& 57.74   & 59.39  & 17.47   & 72.80  & 67.00& 49.87  & 61.71 & 72.21  & 35.18& 8.37   \\ 
(7)& && $\checkmark$  & $\checkmark$ & 27.62   & 75.06  & 28.85   & 86.60  & 76.60& 49.61  & 42.66 & 75.42  & 36.91& 11.96  \\ \midrule
(8)&$\checkmark$&& $\checkmark$  & $\checkmark$& 31.28   & \underline{81.29}  & \textbf{33.21}   & 79.40  & 61.40& 51.70  & 35.12 & 77.54  & 37.66& 13.34  \\
(9)& & $\checkmark$ & $\checkmark$  & $\checkmark$& 61.40   & 74.56  & 32.20   & \underline{87.80}  & \textbf{81.40} & \textbf{57.70}  & \underline{74.40} & \underline{82.98}  & \underline{47.69}& \textbf{14.51}  \\ \midrule
(10)&$\checkmark$& $\checkmark$ & $\checkmark$  & $\checkmark$& 32.45   & \textbf{82.46}  & \underline{32.99}   & 80.40  & 59.20& 52.48  & 34.33 & 78.32  & 37.42& 13.90  \\
\bottomrule
\end{tabular}}
\caption{\textbf{Ablation study of different semantic annotation components on AUTO-UI.} SD and PAR mean \textbf{s}creen \textbf{d}escription and \textbf{p}revious \textbf{a}ction \textbf{r}esult, whereas AT and AD represent \textbf{a}ction \textbf{t}hink and next \textbf{a}ction \textbf{d}escription, respectively. For CLICK and TYPE actions, which is more complicated than the other three, we additionally report the action type prediction accuracy, marked as `type' in this table. Total action-matching score is also included. `GP' is short for goal progress. The best result is marked in \textbf{bold} while the runner-up is \underline{underlined}.}
\label{tab:finetune}
\vspace{-0.2cm}
\end{table*}

\section{Experiments}

\subsection{Zero-Shot Evaluation}

We perform a zero-shot evaluation to investigate the benefit of directly using these screen and action semantics as input. 
Here, we select CogAgent~\cite{hong2023cogagent} for illustration as it is trained to perform GUI tasks and expected to possess generalization abilities since its foundation language model is CogVLM-7B. 
We verify the impact of the proposed chain-of-action thought by adding action think to the prompt input of CogAgent. As shown in Table~\ref{tab:main}, CoAT contributes significant improvements to the overall model performance. Moreover, the first and last line in Table~\ref{tab:main} indicate the fact that fine-tuning a small agent with model size \textasciitilde 1B (i.e. AUTO-UI-base ~\citep{zhan2023you}) using CoAT can obtain comparable performance with a LLM-based agent, demonstrating the strong potential of CoAT on GUI navigation tasks. 

A more detailed comparison between CogAgent and AUTO-UI on model architecture, training data and performance can be found in Appendix~\ref{appendix:comp_baseline}.


\begin{figure}
\setlength{\belowcaptionskip}{0cm}
\setlength{\abovecaptionskip}{-0.1cm}
    \centering
    \includegraphics[width=\linewidth]{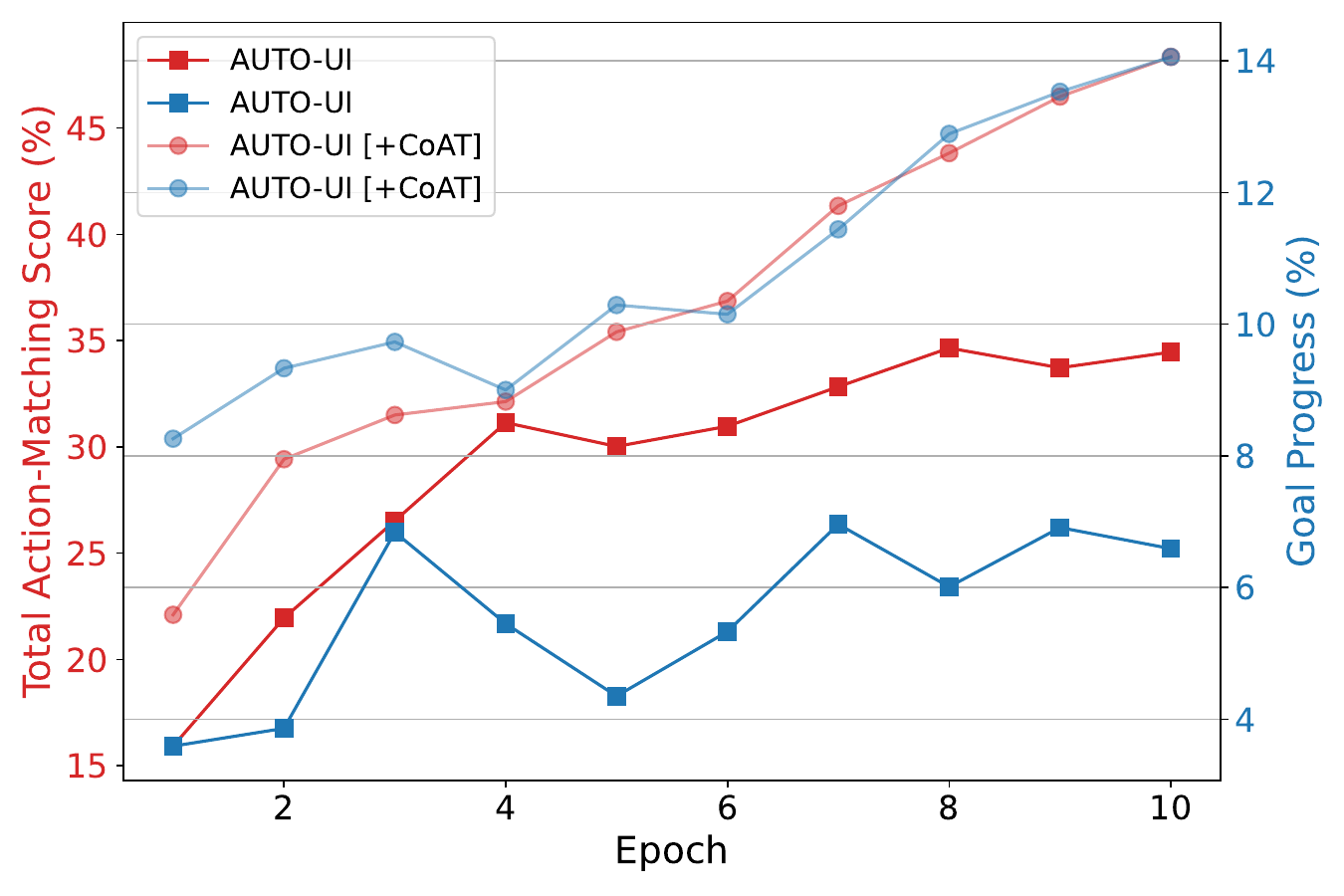}
    \caption{\textbf{Total action-matching score and goal progress over training epochs on AUTO-UI model.}}
    \label{fig:train_log}
    \vspace{-0.3cm}
\end{figure}

\subsection{Fine-tuning Evaluation}

To evaluate the influence of individual components of CoAT, we perform an ablation study by incorporating them alternately. We split the annotations into `input' and `output' groups, indicating where the extra information comes in during the model training. Specifically, we put screen description and previous action result as additional input information, as they do not provide direct help to the current action decision. Action think and next action description are added to the output so that the agent can learn such thinking process.

From Table~\ref{tab:finetune}, we observe that previous action result, especially combined with action think and action description, significantly improve the overall action prediction accuracy of AUTO-UI. As \textbf{the coherence of decision-making process is enhanced by previous action result}, there is a notable increase in the STOP action-matching score (from 60.12 to 79.17). Experiment (5)\textasciitilde (7) demonstrate that \textbf{learning to engage in action thinking without additional input is challenging}. However, when screen description and/or previous action result are added to the input, the performance of AUTO-UI improves immediately, especially in predicting CLICK actions. This validates the necessity and effectiveness of such semantic annotations. There is a minor decrease in both action-matching score and the goal progress when screen description is added, as seen in line (9) and (10). We attribute this to the low resolution of the visual encoder used by AUTO-UI, resulting in an inability to effectively utilize the information in screen descriptions.
Figure~\ref{fig:train_log} further illustrates the improvement in training efficiency when trained with our \textbf{\textsc{AitZ}} data.

\begin{figure*}[t]
\vspace{-0.4cm}
\setlength{\belowcaptionskip}{0cm}
\setlength{\abovecaptionskip}{0.1cm}
    \centering
    \includegraphics[width=\linewidth]{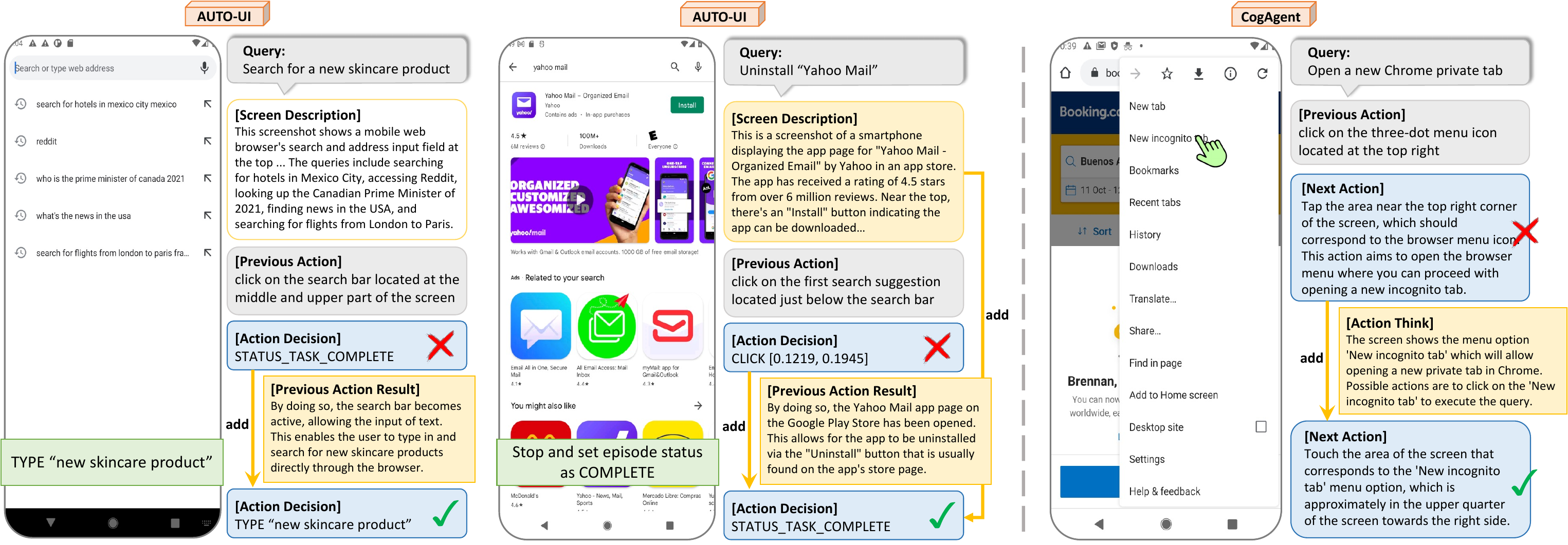}
    \caption{\textbf{Qualitative examples for AUTO-UI and CogAgent.} This figure presents qualitative results where different types of errors are corrected by applying additional semantic annotations (yellow shadowed boxes). }
    \label{fig:case_study}
    \vspace{-0.1cm}
\end{figure*}

\subsection{Qualitative Analysis}
We conduct the thorough analysis on wrong cases, as shown in Figure~\ref{fig:case_study}. 
AUTO-UI struggles with correctly judging the task execution progress, as the action history provided as a series of action types and coordinates is hard to understand. Previous action result mitigates this problem by explicitly describing the result of the previous actions in words. This highlights the importance of safeguarding the coherence of action decision by establishing connections between two time steps. 
For CogAgent, we carefully inspect its output, which is composed of three parts: action plan, next action and grounded operation. It seems that CogAgent does not take historical information into account, as its predictions at each step only consider the current information, leading to repetitive and ineffective actions. 
Adding a short-cut action chain-of-thought, i.e. action think from \textbf{\textsc{AitZ}} dataset, into the model input helps to alleviate this issue.

\section{Related Works}

\paragraph{GUI Navigation} Automatic execution of user instructions on smartphones or websites is an advanced task, as it requires the agent to not only perceive but also deduce. Previous works concentrate on evaluating the ability of models to identify different UI elements ~\citep{shi2017world, zhang2021screen,sunkara2022towards}, and to fulfil a user-queried task by either statically operating on a series of pre-collected GUI screenshots ~\citep{li2020mapping,venkatesh2022ugif,zhan2023you,deng2023mind2web} or dynamically interacting with an alive Android device ~\citep{yang2023appagent}. However, these works separate the ability of element recognition and action inference, causing a discrepancy between the user intent and the performed actions~\citep{wei2022chain, Baechler2024ScreenAIAV}. 
Our CoAT framework bridges this gap by allowing GUI agents to recall history actions, perceive the current screen, and decide on the future actions based on these useful semantics.

\paragraph{Large Multimodal Models (LMM)} 
Recent years have witnessed the rise of numerous large multimodal models~\citep{liu2023improved, liu2023visual,zhu2023minigpt,zeng2023matters}. 
Usually, visual signals are encoded by a vision transformer~\citep{dosovitskiy2020image} and further incorporated in LLMs~\citep{radford2021learning} through linear projection~\citep{tsimpoukelli2021multimodal}, Q-former~\citep{li2023blip} or cross-attention layers~\citep{alayrac2022flamingo}.
For general purpose LMMs, the low resolution of visual encoders (224$\times$224) captures only coarse visual information. 
CogAgent~\citep{hong2023cogagent} deals with this problem by using the original ViT-L~\citep{dosovitskiy2020image} to encode high-resolution visual features up to 1120$\times$1120, and fusing them with every decoder layers through cross-attention. Whereas Monkey~\citep{li2023monkey} equips the visual encoder from QWen-VL ~\citep{bai2023qwen} with individual LoRA adapter ~\citep{hu2021lora} for each patch to scale the image resolution up to 896$\times$1344 pixels. Consequent works~\cite{yu2024texthawk, chen2024far, lu2024deepseek} all incorporate high-resolution image encoders, indicating a popular trend for the future.

\paragraph{LMM as GUI Agents} A number of works have utilized LMMs' domain knowledge and emergent zero-shot embodied abilities to perform complex task planning and reasoning ~\citep{yang2023mm,wang2023describe,ikeuchi2023applying}. For GUI navigation, the introduction of LMMs surpasses previous works that transform the UI layouts and elements into the text-only HTML format ~\citep{li2020interactive, zhang2021screen, wang2023enabling}. One line of work adopts GPT-4V directly as the GUI agent and prompts it to perform the task ~\citep{yan2023gpt, yang2023appagent, zheng2024gpt}, while other methods focus on tuning a smaller LMM on GUI-related datasets to acquire the domain-specific knowledge ~\citep{zhan2023you}, or train a LMM from scratch on GUI-specified pre-training tasks~\cite{hong2023cogagent, baechler2024screenai, you2024ferret, cheng2024seeclick}. 
We evaluate two agents on the proposed \textbf{\textsc{AitZ}} dataset, and prove that our proposed chain-of-action-thought helps agents adapt to GUI tasks better and more quickly.

\section{Conclusion}

In conclusion, our work aims to bolster the navigation ability of LMM-based GUI agents. We propose Chain-of-Action-Thought (\textbf{CoAT}) by analyzing human orienteering processes. We start by verifying that CoAT is superior to three typical context modeling methods. In order to inject CoAT-like thinking capabilities into existing GUI agents, we further generated a set of high-quality CoAT-driven data through cooperation between human experts and GPT-4V, namely Android-In-The-Zoo (\textsc{\textbf{AitZ}}) dataset. \textsc{\textbf{AitZ}} enriches this field with a robust dataset that bridges perception and cognition, facilitating effective training and reliable evaluation for GUI navigation agents. Experiments demonstrate the efficiency and usefulness of proposed chain-of-action-thought paradigm. 

\section{Limitations}

We developed \textbf{CoAT} and \textsc{\textbf{AitZ}} with the goal of enabling LLM Agents to mimic the cognitive processes of humans. Although our experiments proved that it is possible to stimulate the reasoning ability of language models (i.e. GPT-4V~\cite{openai2023gpt4}, CogAgent~\cite{hong2023cogagent} and AUTO-UI~\cite{zhan2023you}) in GUI scenarios through zero-shot prompting or fine-tuning, the different model structure and training data used by current specified models for GUI tasks make the comparison less intuitive. To what extent the image resolution and GUI-related pretraining tasks (i.e. text recognition, GUI imagery~\cite{hong2023cogagent}, screen question-answering~\cite{baechler2024screenai,you2024ferret} and GUI grounding~\cite{cheng2024seeclick}) influence the navigation performance remains under-explored. We leave it for future work to precisely measure the impact of image resolution, text recognition ability, GUI grounding ability of LMMs on GUI navigation tasks.


\section{Ethics}


Android-In-The-Zoo (\textsc{\textbf{AitZ}}) dataset is sourced from open-source datasets AITW~\cite{rawles2023android}, which is permitted for academic use. 
During our data collection, specifically, during the instruction-episode correctness checks, we ensured that privacy concerns were addressed, and the sampled data does not include any real personal information (fake or meaningless data are allowed). Since \textsc{\textbf{AitZ}} dataset contains only semantic annotations on smartphone operations, the use of this data poses neither ethical risks nor harmful guidance.

\bibliography{custom}


\begin{figure*}[t]
\setlength{\belowcaptionskip}{0cm}
\setlength{\abovecaptionskip}{0.1cm}
    \centering
    \includegraphics[width=\linewidth]{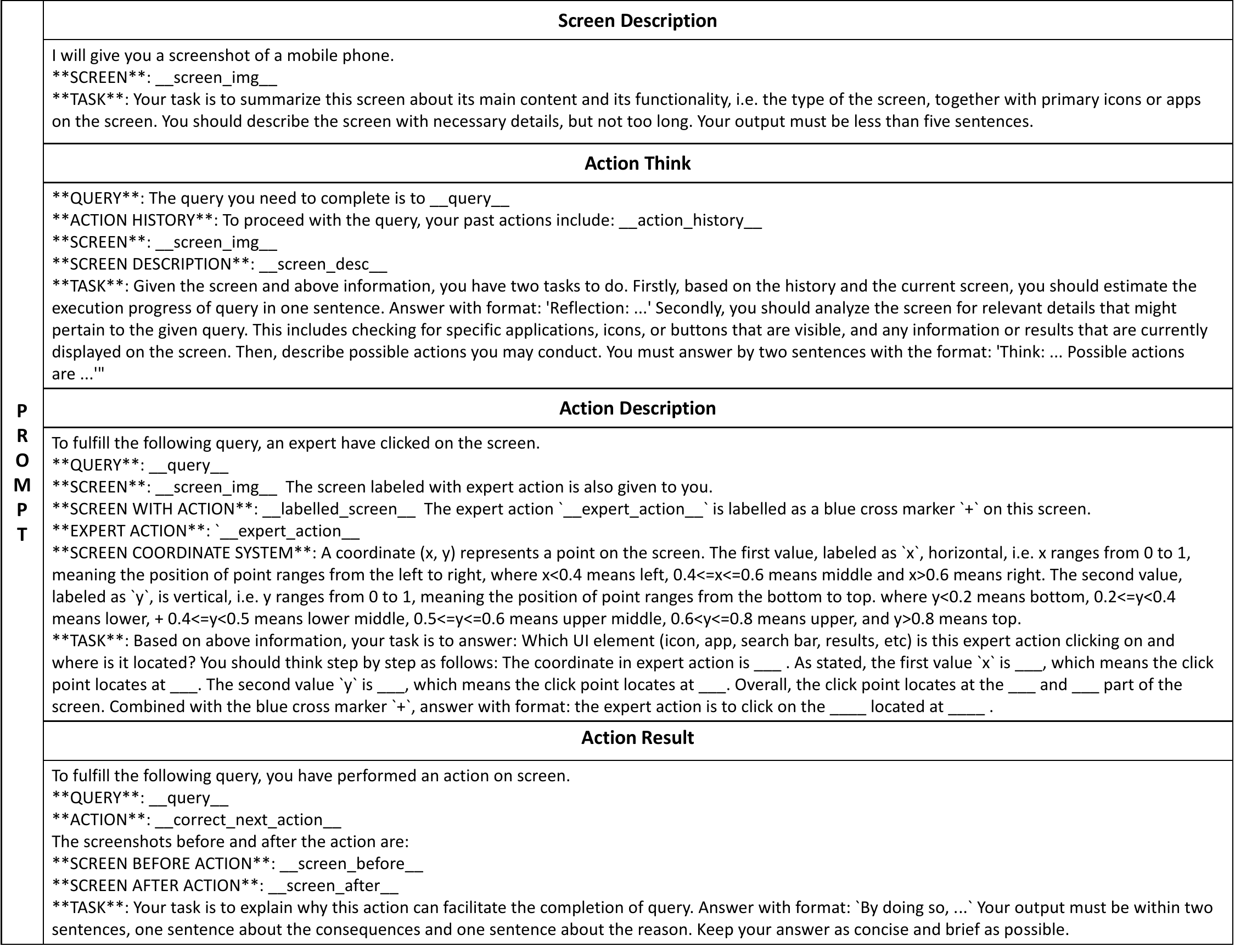}
    \caption{Prompt to generate candidate answers for four types of semantic annotations.}
    \label{fig:prompt}
\end{figure*}

\begin{figure*}
    \centering
    \includegraphics[width=\linewidth]{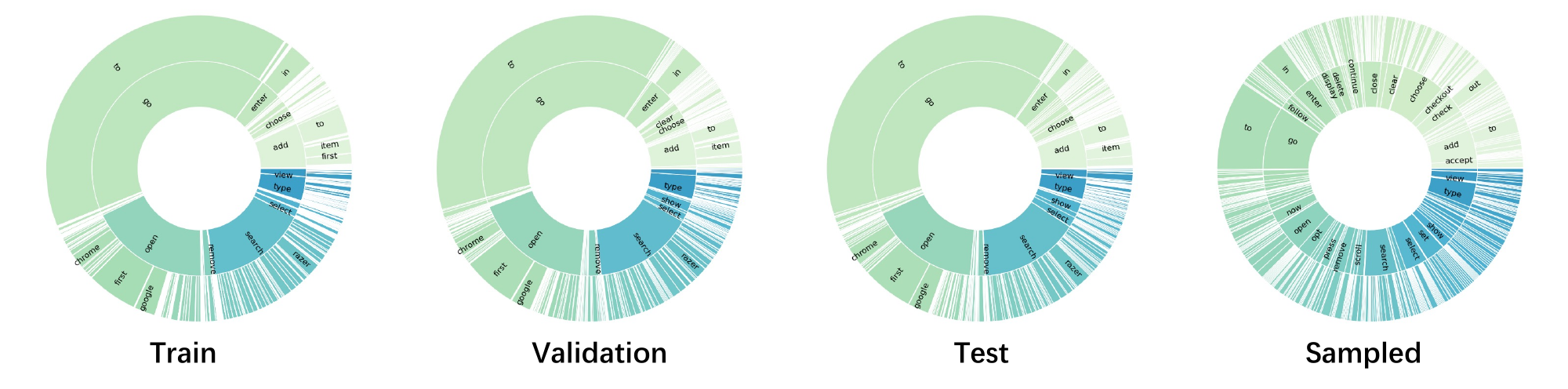}
    \caption{The instruction distribution (grouped by verbs and nouns) for SINGLE subset in original AITW dataset.}
    \label{fig:instr_dist_single}
\end{figure*}

\newpage

\appendix

\section{Data Collection}

Our data construction pipeline is shown in Figure~\ref{fig:data_collection}. We leverage the strong world knowledge and generation ability of GPT-4V, combined with critical human verification, to ensure high-quality data. We ask our annotators to detect only factual errors, which could hardly introduce human bias. 

\subsection{Instruction Sampling Process}

We first checked the instruction distribution in the original AITW dataset based on the split from \cite{zhan2023you}. We find that the instruction distribution in the training, validation and test sets are almost the same, which means there is a serious problem of data leakage. To avoid such problem exists in our constructed datasets, we perform instruction sampling. 
\begin{itemize}[topsep=0pt,leftmargin=*]
    \setlength{\itemsep}{1pt}
    \setlength{\parsep}{2pt}
    \setlength{\parskip}{0pt}
    \item SINGLE: Given the complexity and variety of instructions in this dataset, we first clustered them and then performed balanced sampling based on the categories. The clustering process is as follows: (1) Identify the main verb in each instruction, typically the first word, and group the instructions by this verb. (2) For each group of instructions, we manually classify those with fewer than 50 samples that show clear patterns. Then for groups with more than 50 samples, use tf-idf for clustering. Finally, we manually verify the clustering results. 
    \item WEB\_SHOPPING: We performed balanced sampling based on the types of shopping websites and the objects involved.
    \item GENERAL, INSTALL, GOOGLE\_APPS: Since these three datasets have a limited number of instructions, we did not perform extensive filtering during sampling. Instead, we uniformly sampled x instructions per user. For INSTALL and GENERAL, x=3; for GOOGLE\_APPS, x=5.
\end{itemize}
During the instruction sampling stage, we recruited 10 annotators to verify whether the episodes have successfully completed the tasks required by the instructions. Our data quality inspection team conducted a secondary validation of sampled results.

\subsection{Semantic Annotation Process}
\label{appendix:seman_anno}

We leverage GPT-4V through Amazon Azure-API as the navigation expert and prompt it to do the following generation tasks:
\begin{enumerate}[topsep=0pt,leftmargin=*]
    \setlength{\itemsep}{1pt}
    \setlength{\parsep}{2pt}
    \setlength{\parskip}{0pt}
    \item \textbf{Screen Description}: describe the main content of the given screenshots, including the screen type, and primary apps or widgets presented.
    \item \textbf{Action Grounding}: given the coordinates of the correct next actions, generate action descriptions. Specifically, we simplify the action spaces into 5 action categories, including SCROLL(direction), TYPE(text), PRESS(button), CLICK(point) and STOP(task state). We ask GPT-4V to describe the UI element the click action is operating on, by drawing the bounding box of the clicked area through icon detection model from ~\citep{liu2018learning}. The descriptions for other types of actions are generated using templates.
    \item \textbf{Action Thinking}: think about what actions need to be performed on the current screen to complete the user query, and describe the results of the correct next action based on screenshots before and after the action.
\end{enumerate}
Three experts who have a good understanding of UI elements are recruited as annotators to verify whether the generated action description matches the labelled golden actions and the generated action thinkings. Once inconsistency is found, annotators will manually revise the action descriptions, and enforce GPT-4V to regenerate the action thoughts based on the correct action descriptions. The prompt we use are shown in Figure~\ref{fig:prompt}.

\begin{table*}[t]
\setlength{\abovecaptionskip}{0.15cm}
\setlength{\belowcaptionskip}{0cm}
\centering
\resizebox{1\textwidth}{!}{
\begin{tabular}{c|ccc|cc}
\toprule
  & \multicolumn{3}{c}{\textbf{Model Architecture}} & \multicolumn{2}{|c}{\textbf{Training Data}} \\ \midrule
  & \begin{tabular}[c]{@{}c@{}} Visual \\ Encoder\end{tabular} & \begin{tabular}[c]{@{}c@{}} Language \\ Backbone\end{tabular}&  \begin{tabular}[c]{@{}c@{}} Image \\ Resolution\end{tabular} &  Pre-training & Fine-tuning  \\ 
\midrule
\begin{tabular}[c]{@{}c@{}}\textbf{AUTO-UI} \\ (1.2B)\end{tabular}  & \begin{tabular}[c]{@{}c@{}}Single Encoder (985M) \\ BLIP2-opt-2.7b\end{tabular} & \begin{tabular}[c]{@{}c@{}} FLAN-alphaca \\ -base(200M)\end{tabular}& 224 x 224   &  /    & AITW / \textbf{\textsc{AitZ}}   \\
\midrule
\begin{tabular}[c]{@{}c@{}}\textbf{CogAgent} \\ (18B)\end{tabular}
& \begin{tabular}[c]{@{}c@{}}Dual Encoder (11B)\\ Low-Res: EVA2-CLIP-E \\ High-Res: EVA2-CLIP-L\end{tabular}   & CogVLM-7B  & 1120 x 1120   &  \begin{tabular}[c]{@{}c@{}}276M data  spanning \\ over text recognition, \\ visual grounding \\ and gui imagery tasks\end{tabular} & \begin{tabular}[c]{@{}c@{}}1M data, including \\  Mind2Web, AITW, \\ public VQA data ...\end{tabular} \\ 
\bottomrule
\end{tabular}}
\caption{\textbf{Comparison between AUTO-UI~\cite{zhan2023you} and CogAgent~\cite{hong2023cogagent}.} Note that, for low-resolution images, following CogVLM~\cite{Wang2023CogVLMVE}, CogAgent adopts a visual encoder with 5B parameters and a visual expert module with 6B parameters. The high-resolution encoder only takes 0.3B parameters.}
\label{tab:cmp_autoio_cogagent}
\end{table*}

\subsection{Action Space}
\label{appendix:action_space}

As stated before in Appendix~\ref{appendix:seman_anno}, we simplify the action spaces into 5 action categories. The reason behind this is, we observe that within the AITW dataset, `DUAL\_POINT' action type seamlessly covers both `CLICK' and `SCROLL' actions. In most cases, the action point of `SCROLL' action conveys little information, but the scroll direction matters. There are also few operations that require dragging apps, such as editing the main screen. Therefore, we manually split the `DUAL\_POINT' action type into `CLICK' and `SCROLL', where `CLICK' action involves coordinate prediction and `SCROLL' action is purely textual. The action space is summarized as follows:
\begin{itemize}[topsep=0pt,leftmargin=*]
    \setlength{\itemsep}{1pt}
    \setlength{\parsep}{2pt}
    \setlength{\parskip}{0pt}
    \item \textbf{\texttt{CLICK(coord\_y: float, coord\_x: float)}}: 
    This action clicks a specific point on the screen. It is necessary to combine the annotation of UI elements to identify the icon and/or area clicked. Note that we use the relative pixel coordinate system, where (0, 0) means the top-left and (1,1) means the bottom right corner of the screen. For example, \texttt{click} \texttt{(0.11, 0.92)} taps a point located at the top-right corner of the screen.
    \item \textbf{\texttt{SCROLL(direction: str)}}: 
    This actions means the finger movements like a real human user. For example, \texttt{scroll up} means the action gesture is from bottom to top, leading either the app drawer to be opened, or the current screen to go down and reveal more contents.
    There are four options for direction: \texttt{up}, \texttt{down}, \texttt{left} and \texttt{right}. 
    \item \textbf{\texttt{TYPE(text: str)}}: 
    This action allow the agent to directly type texts into an input field, skipping the inefficient keyboard operations. For example, \texttt{type ``what is CoAT''} inputs the string ``what is CoAT'' to the text input field at one time.
    \item \textbf{\texttt{PRESS(button: str)}}: 
    The Android system provides several system level shortcut buttons, such as \texttt{back button} that enables the user back to the previous interface, and \texttt{home button} that allows a direct return to the home screen. Moreover, \texttt{enter button} is another virtual button that submits the typed query. This action means to press on one of the system level virtual buttons.
    \item \textbf{\texttt{STOP(task\_state: str)}}: 
    This action allows the agent to stop and end the query execution in time, either when it considers the task is completed or the task is impossible. 
    For example, \texttt{stop and set the query as completed} means the user query has been successfully completed. 
\end{itemize}
We map the actions predicted by AUTO-UI and CogAgent to this space to ensure the reliability and consistency in comparison.

\begin{table*}[t]
\setlength{\abovecaptionskip}{0.15cm}
\setlength{\belowcaptionskip}{0cm}
\centering
\resizebox{1\textwidth}{!}{
\begin{tabular}{c|c|c|c|c|ccccc}
\toprule
Model             & Prompt            & UI Reps. & Hit Rate & Total & CLICK & SCROLL & PRESS & TYPE & STOP \\ \midrule
\multirow{6}{*}{QWen-VL} & \multirow{2}{*}{CoA}  & txt          & 82.53   & 35.86    & 44.96    & 34.21     & 0        & 34.04   & 4.08    \\ 
                        &                       & tag          & 94.48   & 44.37    & 60.07    & 7.89      & 0        & 48.94   & 0       \\ \cmidrule{2-10}
                        & \multirow{2}{*}{CoT}  & txt          & 84.37   & 41.61    & 56.83    & 2.63      & 4.35     & 40.43   & 4.08    \\
                        &                       & tag          & 95.63   & 49.43    & 69.42    & 2.63      & 4.35     & 40.43   & 2.04    \\ \cmidrule{2-10}
                        & \multirow{2}{*}{CoAT} & txt          & 94.02   & 52.41    & 72.3     & 7.89      & 13.04    & 34.04   & 10.2    \\
                        &                       & tag          & 96.32   & 51.95    & 70.5     & 2.63      & 8.7      & 46.81   & 10.2   \\ \midrule
\multirow{6}{*}{Gemini-PV} & \multirow{2}{*}{CoA}  & txt          & 89.43   & 42.99    & 60.79    & 13.16     & 4.35     & 21.28   & 4.08    \\
                        &                       & tag          & 99.77   & 54.48    & 79.86    & 10.53     & 13.04    & 10.64   & 6.12    \\ \cmidrule{2-10}
                        & \multirow{2}{*}{CoT}  & txt          & 95.86   & 49.2     & 67.27    & 26.32     & 21.74    & 19.15   & 6.12    \\
                        &                       & tag          & 97.47   & 51.95    & 74.46    & 21.05     & 13.04    & 12.77   & 4.08    \\ \cmidrule{2-10}
                        & \multirow{2}{*}{CoAT} & txt          & 97.01   & 52.41    & 69.42    & 23.68     & 30.43    & 34.04   & 6.12    \\
                        &                       & tag          & 95.4    & 53.33    & 72.66    & 23.68     & 21.74    & 29.79   & 4.08    \\ \midrule
\multirow{6}{*}{GPT-4V} & \multirow{2}{*}{CoA}  & txt          & 92.41   & 55.17    & 74.1     & 42.11     & 39.13    & 8.51    & 10.2    \\
                        &                       & tag          & 99.31   & 62.76    & 86.69    & 44.74     & 26.09    & 14.89   & 4.08    \\ \cmidrule{2-10}
                        & \multirow{2}{*}{CoT}  & txt          & 98.16   & 66.21    & 89.57    & 39.47     & 39.13    & 12.77   & 18.37   \\
                        &                       & tag          & 97.01   & 64.14    & 86.33    & 39.47     & 39.13    & 21.28   & 10.2    \\ \cmidrule{2-10}
                        & \multirow{2}{*}{CoAT} & txt          & 98.39   & 71.72    & 86.33    & 47.37     & 43.48    & 48.94   & 42.86   \\
                        &                       & tag          & 98.16   & 71.49    & 86.69    & 42.11     & 43.48    & 57.45   & 34.69   \\ \bottomrule
\end{tabular}}
\caption{\textbf{Complete comparison results of three prompting methods} on Qwen-VL-Max, Gemini-1.0-Pro-Vision and GPT-4V. ``Prompt'' means different prompting methods. ``UI Reps.'' denotes the representation methods of screen elements, including set-of-mark tagging (tag) and textual representation (txt). ``Hit Rate'' means the format hit rate. The evaluation metric is the action prediction accuracy(\%).}
\label{tab:full_cmp}
\end{table*}

\section{Experiment Details}

\subsection{Comparison between Prompting Methods}
\label{appendix:comp_prompt}


In Section~\ref{sec:cmp_prompt} we conducted a preliminary experiment to demonstrate that CoAT is more effective than previous context modeling methods. Specifically, for CoA prompting, the input to GUI agents includes system prompt, current screenshot, history actions and user request. For CoT prompting, the input to GUI agents includes system prompt, current screenshot and user request. For CoAT promprting, we firstly require the agent to observe current screenshot and generate screen descriptions. Then, the input contains system prompt, current screenshot, screen description, history actions, previous action results and user request. 

\begin{figure*}[t]
\setlength{\belowcaptionskip}{0cm}
\setlength{\abovecaptionskip}{0.1cm}
    \centering
    \includegraphics[width=\linewidth]{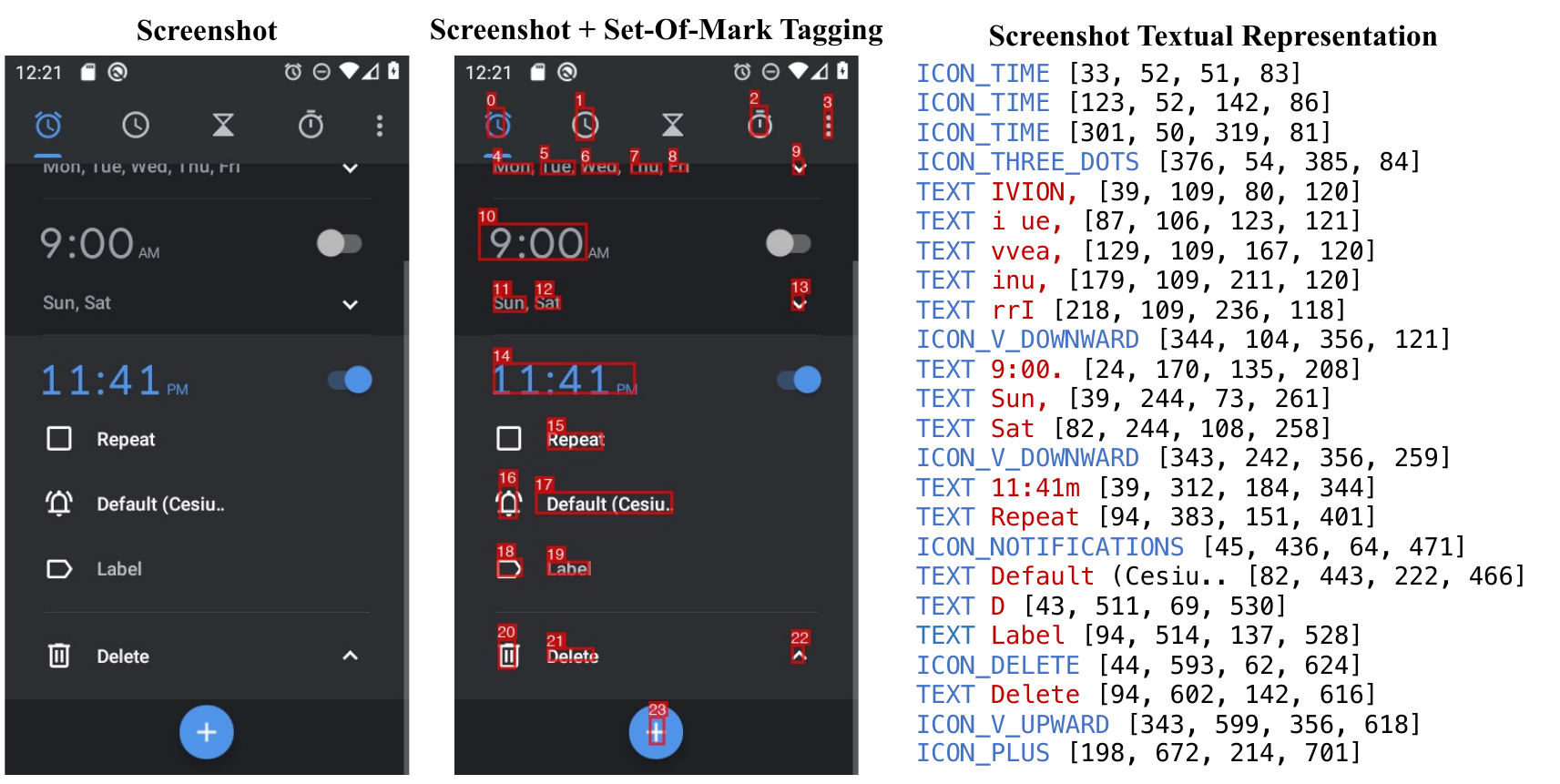}
    \caption{\textbf{Visualization of Set-Of-Mark tagging and corresponding textual representations.}}
    \label{fig:screen_reps}
    \vspace{-0.0cm}
\end{figure*}

For all threee prompting methods, the system prompt contains information about the valid action space and corresponding desired output format. If the representation of UI elements is set-of-mark tagging, another screenshot with annotated UI elements will be additionally added to the input, otherwise a textual representation of UI elements is appended. Figure~\ref{fig:screen_reps} show a visualization example of these two UI representations.

The complete experiment results are shown in Table~\ref{tab:full_cmp}. From Table~\ref{tab:full_cmp}, GPT-4V prompted by CoAT takes the lead position in the overall performance and in the prediction of each type of actions. 
Compared with plain textual representations, agents equipped with set-of-mark tagging generally performs better. This encourages future work to put more emphasis on the visual perception of UI elements, improve the image resolution and multi-image processing ability of GUI agents.

\subsection{Comparison between Baselines}
\label{appendix:comp_baseline}

As shown in Table~\ref{tab:main}, we conclude that ``Auto-UI + CoAT is on par with CogAgent-Chat-18B'' based on the fact that the model architecture and training data of AUTO-UI is inferior to CogAgent, but after fine-tuning on \textbf{\textsc{AitZ}} dataset, they achieve similar performance on goal process (AUTO-UI is even slightly higher.) We summarize the differences between two models in Table~\ref{tab:cmp_autoio_cogagent} for a quick look. Following is the detailed explanation:

\begin{enumerate}[topsep=0pt,leftmargin=*]
    \setlength{\itemsep}{1pt}
    \setlength{\parsep}{2pt}
    \setlength{\parskip}{0pt}
    \item The lesser volume of training data used by Auto-UI compared with CogAgent. Specifically, Auto-UI underwent fine-tuning solely on the \textbf{\textsc{AitZ}} dataset, in contrast to CogAgent's extensive fine-tuning across the entire AITW dataset. Moreover, CogAgent introduced GUI imagery tasks during the pre-training phase. Hence, it is highly optimized for GUI scenarios. 
    \item The different resolution of visual encoders. Specifically, Auto-UI employs the visual encoder from BLIP2 with a 224 x 224 resolution, whereas CogAgent combines ViT-L with the visual encoder from CogVLM to scale the resolution up to 1120 x 1120. 
    \item Despite Auto-UI + CoAT being trained with significantly less data and without any additional pre-training efforts, it managed to outperform CogAgent in terms of action prediction accuracy and goal progress, underscoring the effectiveness and value of our proposed method and dataset, as shown in Table~\ref{tab:main}.
\end{enumerate}

\end{document}